# CORE: A Conceptual Reasoning Layer for Large Language Models[1]

Vishwas Hegde  vishwashegde@gmail.com
Vindhya Shigehalli  vindhya.shigehalli@gmail.com
Independent Researchers

## Abstract

Large language models handle single-turn generation well, but multi-turn interactions still require the model to reconstruct user intent and task state from an expanding token history because internal representations do not persist across turns. This token-first paradigm leads to drift, inconsistent reasoning modes, and growing prompts as conversations deepen. We propose CORE, a concept-first interaction layer that improves multi-turn stability without modifying model weights. CORE combines a small library of universal cognitive operators with a persistent Local Concept—a compact semantic state capturing the task, constraints, preferences, and intermediate results. Each model call receives only this concept state, the user's latest instruction, and the selected operator, eliminating the need to replay full history. A preliminary prototype simulating CORE's behavior shows a ~42% reduction in cumulative prompt tokens, though this number reflects prototype conditions and should not be interpreted as a real-world performance estimate. CORE offers a model-agnostic mechanism that separates conceptual reasoning from language generation, suggesting a scalable direction for more stable multi-turn systems.

## 1  Introduction

Large language models (LLMs) have achieved remarkable fluency, generality, and problem-solving capability. However, despite their expressive power, today's LLM systems operate within a fundamentally token-first interaction paradigm. Each model invocation receives a sequence of tokens, produces a sequence of tokens, and then discards its internal representations. The system does not retain a persistent understanding of the task being performed. Because meaning exists only in transient hidden states, multi-turn dialogue requires the model to repeatedly infer user intent, extract constraints, and reconstruct intermediate reasoning solely from the token history that fits within the current context window.

This architecture works surprisingly well in short exchanges, yet it creates structural limitations as conversations become longer, more complex, or more task-oriented. The model must continually reinterpret earlier turns, often under partial visibility when the context window truncates past interactions (Vaswani et al., 2017). This leads to redundant computation, unstable reasoning modes, inconsistent task structure, and a well-documented tendency for responses to drift away from the user's original objectives (Bang et al., 2023). Over time, the prompt grows linearly as the system replays previous turns, and the model alternates between reconstructing meaning and generating new content.

Human cognition operates very differently. Rather than rebuilding meaning from the ground up on each turn, humans rely on conceptual structures that remain stable across different linguistic expressions. Piaget's theory of cognitive development describes these structures as schemas, which are organized frameworks that help individuals interpret and organize incoming information. As people encounter new experiences, schemas evolve through assimilation, which incorporates new information into existing frameworks, and accommodation, which adjusts existing frameworks when the information does not fit (Piaget and Inhelder, 1969; The Development of Children, 2018). These processes allow humans to maintain continuity of meaning and to update their understanding in a coherent and structured way.

The idea of concept-level structure in artificial intelligence is not new. Early symbolic systems relied on handcrafted ontologies that encoded classes and relations, but these systems were rigid, manually engineered, and unable to expand beyond the categories defined by human designers. Modern language models, however, learn relational and semantic regularities directly from data. Recent work shows that pretrained language models can perform tasks such as knowledge graph completion, which indicates that they capture structured associations that resemble conceptual relationships (Yao et al., 2019). This suggests an opportunity to revisit conceptual layers as dynamic structures that can be inferred, extended, and updated by contemporary models.

The central hypothesis of this paper is that these limitations arise not from deficiencies of the underlying model, but from the absence of a concept-level interaction layer. If the system had access to a persistent representation of the task—separate from the ever-growing token transcript—then multi-turn reasoning could become more stable, more efficient, and more interpretable. Such a layer would not replace model training or modify internal weights; rather, it would sit alongside the model as a complementary mechanism that governs how meaning is carried forward across turns.

To explore this idea, we propose **CORE**, a concept-first reasoning layer that externalizes two aspects of multi-turn interaction that are currently implicit. The first is the structure of reasoning itself, represented as a small, fixed set of cognitive operators that describe how the model should transform meaning on a given turn. The second is a Local Concept state that captures the evolving semantics of the task—including goals, constraints, preferences, and intermediate outputs—and persists across turns independently of surface tokens. By separating the representation of meaning from the representation of text, CORE enables systems to construct prompts that are grounded not in replayed conversation snippets, but in a compact, stable concept-level summary.

This shift from token-first to concept-first interaction does not require architectural changes to the underlying transformer. Instead, it reframes the interface between users and models, offering a simple but powerful mechanism for improving the consistency and scalability of multi-turn systems. The remainder of this paper develops this architecture, contrasts it with existing token-based approaches, and presents an early prototype demonstrating its potential.

## 2   Related Work

Prior efforts to improve LLM stability and structure span several directions, each addressing specific shortcomings of token-based interaction. CORE intersects with these lines of work but diverges in its objective: providing a persistent, concept-level representation that survives across model invocations.

**Prompt engineering** improves single-turn reliability by shaping how models interpret instructions, but it operates only at the level of textual cues (Bender et al., 2021; Ouyang et al., 2022). It creates no persistent semantic state and offers no mechanism for stabilizing reasoning across turns. CORE likewise emits structured cues, but they originate from an explicit meaning-level state rather than ad hoc prompt manipulation.

**Chain-of-thought and related prompting methods** enhance within-invocation reasoning but do not carry intermediate understanding forward (Wei et al., 2022; Press et al. 2023). Explanations produced in one call are consumed as text in the next and do not form a durable substrate. CORE is compatible with such techniques yet targets the orthogonal problem of cross-invocation continuity.

**Memory-augmented and retrieval-based systems** expand contextual grounding by surfacing external documents or embeddings (Lewis et al., 2020). These methods store information about the world, not the state of the task: user intent, constraints, partial results, or the current reasoning mode. CORE does not store external knowledge; it stores the evolving semantic state of the conversation, which is independent of retrieval mechanisms.

**Agent-based frameworks** maintain plans, tools, or control flows but do not define a domain-agnostic representation of task semantics (Schick et al., 2023). Their state machines encode behavior, not meaning. CORE complements agents by supplying a universal reasoning layer that decouples conceptual state from textual dialogue.

Taken together, existing approaches improve individual capabilities—prompt fidelity, factual grounding, intermediate reasoning, or tool use—but none directly addresses the structural absence of a persistent conceptual interface between the model and the task. CORE targets precisely this gap.

## 2.1 Comparison with Procedural Reasoning Frameworks

Methods such as chain-of-thought, self-ask, tree-of-thought search, and iterative reflection orchestrate multi-step reasoning by coordinating multiple model calls (Wei et al., 2022; Press et al., 2023). They improve the trajectory of reasoning within each invocation but retain the limitations of token-based continuity. Intermediate steps exist only as text; once they fall out of the context window, the system must reconstruct meaning from scratch.

CORE addresses a different structural concern. Instead of relying on textual replay, it introduces a non-textual, persistent semantic substrate—the Local Concept—updated explicitly each turn and surfaced to the model as a compact conceptual summary.

A further distinction lies in reasoning control. Procedural prompts encourage open-ended processes ("think step by step," "explore alternatives"), but they do not define a stable, domain-agnostic grammar of reasoning. CORE instead imposes a finite library of cognitive operators that standardize reasoning transformations across tasks.

Finally, procedural methods do not prevent drift across turns: a well-shaped chain-of-thought on one call does not constrain the next call's reasoning mode or ensure retention of prior constraints. CORE enforces continuity by grounding every invocation in a persistent Local Concept and an explicitly selected operator.

In this sense, procedural frameworks and CORE occupy complementary layers: procedural prompts optimize within-invocation reasoning, whereas CORE provides the cross-invocation semantic stability that enables coherent multi-turn interaction.

## 2.2 Limitations of Token-Based Multi-Turn Reasoning

Modern language models process text as sequences of tokens. During each forward pass, they build rich internal representations—semantics, intent, context—but these vanish immediately after generation. No conceptual state carries across turns (Vaswani et al., 2017).

As a result, multi-turn dialogue depends entirely on replay: every invocation must reconstruct meaning from the visible transcript. The model repeatedly re-infers the user's goal, constraints, and partial plans, even when nothing has changed. As conversations grow, this reconstruction becomes incomplete as older turns fall out of context and inefficient as the full history must be reprocessed through attention (Bang et al., 2023).

This token-first paradigm produces predictable failures:
- Reasoning drift from re-deriving the reasoning strategy each turn.
- Constraint erosion because tone, audience, and requirements exist only as implicit text.
- Inconsistent structure, since lists, plans, or frameworks have no persistent representation.
- Prompt inflation, where long transcripts are replayed just to maintain coherence.

These issues are inherently architectural: today's systems lack a layer that separates meaning from tokens. Without a persistent conceptual substrate, the model must rediscover the task at every step.

CORE fills this gap by externalizing a stable semantic state so each turn builds on meaning—not on repeated token reconstruction.

## 3 Model Architecture

### 3.1 Global CORE: A Finite Library of Universal Cognitive Operators

CORE begins by formalizing the observation that, while the topics encountered in dialogue are unbounded, the operations used to reason about them are comparatively few. Tasks as varied as choosing a dog breed, analyzing a scientific claim, rewriting an email, constructing a plan, or comparing investment options all draw from a familiar set of cognitive transformations: summarizing, explaining, comparing, decomposing, evaluating, and so forth. In current LLM interfaces, these transformations are inferred implicitly from the token sequence and may vary unpredictably from turn to turn.

Global CORE makes these reasoning moves explicit. It consists of a deliberately bounded library of approximately forty cognitive operators that represent reusable patterns of reasoning. Each operator specifies a mode of transformation—such as "summarize this content," "compare these alternatives," or "decompose this goal into actionable steps"—together with the structural expectations for the model's output. These operators do not encode domain knowledge; they encode how reasoning should proceed rather than what the reasoning is about.

The size of the Global CORE is intentionally limited. Expanding the library to hundreds or thousands of operators would undermine interpretability, invite redundancy, and reintroduce the unbounded complexity that CORE is designed to avoid. Conversely, limiting the library to only a handful of operators would force them to become overly generic, diluting their descriptive power and leading to ambiguity in operator selection. A mid-sized library of roughly forty operators provides sufficient expressiveness to cover the vast majority of conversational tasks while remaining compact enough to ensure transparency and stability.

A complete enumeration of the Global CORE operators, together with brief descriptions of their intended function and output structure, is provided in Appendix A.

These operators constitute a reasoning grammar. Just as a finite set of syntactic rules supports an unbounded variety of natural language expressions, the Global CORE enables a wide range of conversational tasks to be represented through combinations of a small number of reasoning modes. The library is designed to remain static during deployment. While future research may explore data-driven refinement or expansion of operator families, the architecture presented here assumes a stable, fixed operator set that serves as the foundation for structured multi-turn reasoning.

Figure 1 presents a high-level architecture of CORE, showing how user input, operator selection, Local Concept updates, and the conceptual communication packet interact with an unchanged large language model.

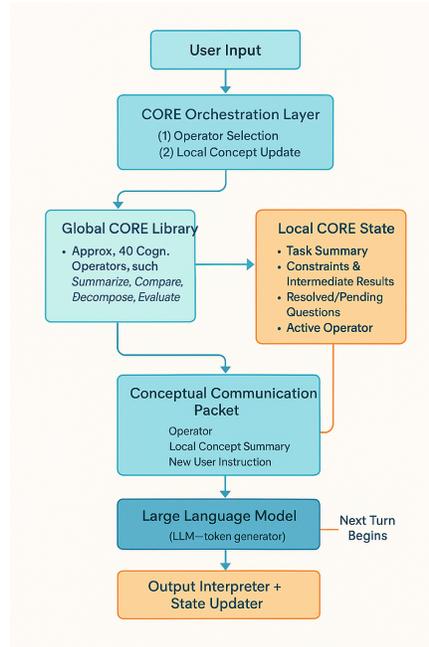

Figure 1: CORE - Model Architecture

## 3.2 Local CORE: The Per-Conversation Concept State

If Global CORE specifies the set of cognitive operations that structure reasoning, Local CORE represents the semantic state of an ongoing task. While the token transcript captures what has been said, the Local Concept captures what the system understands the task to mean. This state persists across turns and provides the model with continuity independent of raw textual history.

### 3.2.1 Structure of a Local Concept

A Local Concept is a compact, structured object consisting of the following fields:

- **Task Summary:** A brief description of the task the user is engaged in (e.g., "selecting an appropriate dog breed for a family with children").

- **Constraints and Preferences:** Conditions introduced explicitly or implicitly by the user (e.g., "low shedding," "hypoallergenic," "apartment-friendly," "formal tone").

- **Intermediate Results:** Conceptual outputs computed in earlier turns and needed for later reasoning (e.g., a shortlist of candidates, an outline, a partially constructed plan).

- **Resolved and Pending Questions:** A record of what ambiguities have already been clarified and what remains unresolved.

- **Active Operators:** The Global CORE operator(s) used in previous steps and the operator selected for the next turn.

These elements form a semantic representation that can be summarized in tens of tokens. The Local Concept does not store text from the conversation; instead, it stores meaning extracted from it.

A sample abstract schema (purely conceptual) is:

```
LocalConcept {
  task_summary: string
  constraints: { key: value }
  intermediate_results: { key: value }
  resolved_questions: [string]
  pending_questions: [string]
  active_operator: OperatorID
  last_updated: timestamp
}
```

This structured format makes Local CORE lightweight, interpretable, and easy to persist across turns.

### 3.2.2 Where Local CORE Is Stored

Local CORE is a system-level construct, external to the model. It is not part of the transformer, its parameters, or its hidden state. Instead, Local Concepts reside in the application or orchestration layer that manages interactions with the model. This design ensures that CORE can be applied to any existing model without altering its architecture.

Three types of storage are possible, depending on deployment:

1. **In-Session Storage**
   For active conversations, Local Concepts are held in a transient session store (e.g., in-memory cache, server-side state). This enables fast access and updates on every turn.

2. **Warm Retention Storage (User-Scoped)**
   To support returning to earlier topics—even days later—the system may persist a bounded number of Local Concepts per user in a lightweight key–value or document store. This storage holds only structured semantic fields, not text. Concepts are retained for a configurable interval (e.g., thirty days), subject to capacity and eviction policies.

3. **Application Layer Responsibility**
   Local CORE does not require any changes to the LLM's architecture or training. It is part of the interface layer that governs how the model is prompted. Different products can choose different retention strategies:

- stateless (delete all Local CORE per session),
- session-based (retain only during the chat),
- profile-based (retain for a fixed time window),
  depending on privacy and product needs.

This separation ensures that Local Concepts remain small, bounded, and implementation-agnostic. They neither accumulate indefinitely nor burden the underlying model with memory they are not designed to carry.

### 3.2.3 Lifecycle Management

To avoid unbounded growth, Local Concepts follow a controlled lifecycle:

- A concept is *active* when used during the current turn.
- It becomes *dormant* when the user shifts topics but remains retrievable from the warm store.
- It is *purged* when its retention window expires or the user's concept capacity limit is reached.

Because Local Concepts store meaning rather than transcripts, even a multi-week retention policy imposes negligible storage cost.

## 3.3 The CORE Interaction Loop

The interaction loop in CORE is not a sequence of prompt manipulations but an explicit state–transition system governing how meaning is represented, updated, and communicated to the model across turns. Its purpose is to ensure that the model operates on a stable conceptual substrate rather than repeatedly reconstructing intent from textual history.

Each turn begins by interpreting the user's new input in the context of the existing Local Concept. From this, the system identifies the appropriate Global CORE operator—that is, the reasoning transformation that should govern the turn. Operator selection defines a formal reasoning mode, determining whether the system is engaged in comparing, explaining, summarizing, planning, decomposing, or another well-specified cognitive operation.

Once the operator is known, the Local Concept is updated. This update is an evolution of the structured semantic state that persists independently of text. New constraints are incorporated, intermediate results are recorded, clarifications are resolved, and task summaries are refined. The Local Concept thereby accumulates continuity of meaning in a way that hidden states cannot and prompts do not attempt to represent.

The system then constructs what is best understood as a conceptual communication packet for the language model. Rather than replaying the token history, the model receives a compact representation of the task's meaning along with the operator that specifies the form of reasoning required for the next step. This conceptual packet enables the model to perform targeted symbolic transformation rather than implicit rediscovery of context.

The model's output is then interpreted as a refinement of the meaning-level state. Any structured information—lists, choices, arguments, resolved ambiguities—is fed back into the Local Concept. In this way, the model participates in a process of semantic state construction rather than ad-hoc textual elaboration.

This loop transforms multi-turn dialogue from a sequence of isolated token-generation events into a coherent stateful reasoning process, with explicit semantics, explicit operations, and explicit memory. It is not an extension of prompting practice but a redefinition of how language models interface with tasks.

## 3.4 Topic Switching and Resumption

Multi-turn conversations rarely proceed linearly. Users shift between topics, suspend tasks, introduce digressions, and return to earlier objectives after minutes, hours, or even days. In token-based systems, such transitions pose challenges: the model must either retain the entire transcript—leading to growing prompts and repeated re-interpretation—or lose access to earlier turns when the context window truncates past interactions. As a result, topic boundaries blur, constraints decay, and conversations often require the user to restate information the model has already seen.

CORE resolves these issues by treating each coherent task or topic as its own Local Concept, allowing the system to manage multiple semantic threads without conflating them or depending on the visibility of old tokens. When the user shifts to a new topic, the currently active Local Concept transitions to a dormant state, and a new Local Concept is initialized for the new topic. Because these concepts store distilled meaning rather than text, they occupy negligible space and can be maintained concurrently without significant memory cost.

When the user later produces input that aligns semantically with a dormant concept—for example, returning to a discussion about dog breeds after a detour into physics—the system reactivates the corresponding Local Concept. The model is then provided with the concept-level summary of that task, allowing it to resume reasoning from the point at which the conversation left off. This does not require replaying the earlier transcript or relying on the model's memory of prior generations. The continuity arises from the persisted semantic state rather than from reconstructed text.

To ensure scalability and prevent unbounded accumulation of dormant topics, the warm store that maintains these concepts is strictly bounded. Dormant concepts are retained for a configurable window (e.g., thirty days) and are evicted when they exceed capacity or become stale, following a least-recently-used policy. Since each concept consists only of structured semantic fields and not conversational text, even multi-week retention imposes minimal overhead.

This mechanism enables natural conversational behavior: users may branch, revisit, and interleave topics without explicitly reminding the system of earlier context. CORE preserves the coherence of each thread by maintaining a persistent, meaning-level representation of its state. The result is a form of conversational continuity that does not depend on token history or context length but on explicit conceptual structure.

## 3.5 Clarifying What CORE Is Not

Although CORE influences how prompts are structured, it should not be understood as an advanced form of prompt engineering. Prompt engineering modifies surface text on a turn-by-turn basis but does not create or maintain a persistent semantic state. In contrast, CORE introduces an explicit, durable representation of task meaning—along with a fixed grammar of reasoning operators—that governs how the system interprets user input and how the model is instructed on each turn. This representation exists outside the model and is preserved independently of token history, enabling stability that cannot be achieved through prompt manipulation alone.

CORE also does not require the model to grasp these representations symbolically. Transformers respond reliably to structured textual signals, and CORE translates the concept state into precisely such signals. The model remains a token generator, but the tokens it receives encode the task state and the applicable reasoning mode explicitly. In this way, CORE constrains multi-turn reasoning not by crafting clever text but by supplying a stable, meaning-level substrate that the model conditions on across turns.

## 3.6 Architectural Rationale

CORE's effectiveness stems from reorganizing how meaning is preserved across turns. Current LLM interfaces rely on token sequences as both communication channel and state store. Because internal representations vanish after each response, systems reconstruct intent and constraints by replaying the transcript, a process that introduces drift and inconsistency as conversations grow.

CORE instead separates meaning from text. An instructive analogy comes from natural language: human languages achieve unbounded expressivity not through infinite rules but through a finite grammar applied to an open vocabulary. The generative power lies in a small, stable set of compositional operations. CORE adopts this structure for reasoning. A fixed library of cognitive operators serves as a "grammar," while the Local Concept provides the evolving "content." This allows an unlimited task space without expanding the set of reasoning moves.

The Local Concept maintains a persistent semantic state, replacing reconstruction with direct access to distilled meaning. Each invocation receives this state along with the selected operator, which specifies the reasoning mode—comparison, planning, explanation, and so on. Transformers need no symbolic understanding; they respond more predictably to short, structured prompts. By regularizing the input distribution in this way, CORE stabilizes multi-turn behavior without altering model weights.

Thus multi-turn dialogue becomes incremental rather than rediscovered: reasoning proceeds by updating a persistent conceptual substrate rather than reinterpreting prior text. This shift underlies the improvements in stability, interpretability, and efficiency explored next.

## 3.7 Efficiency and Stability Dynamics

MORE changes the dynamics of multi-turn interaction by altering what information the model must process and how consistently it processes it. In conventional token-based systems, every turn requires the model to reconstruct user intent and task structure from the visible transcript. As conversations grow, the model reprocesses increasingly long histories, and earlier constraints attenuate or disappear as they fall outside the context window. This reconstruction burden leads to both rising token costs and unstable reasoning trajectories.

By externalizing meaning into the Local Concept, CORE removes the dependence on replay. The prompt for each turn contains a compact summary of the task rather than an accumulated transcript, so prompt size approaches a constant bound instead of growing linearly. This reduces redundant computation and mitigates the prompt-expansion costs that dominate long conversations.

Stability improves for a different reason. In token-first systems, the model infers its reasoning mode from context, and this inference can drift as the dialogue unfolds: a comparison may slowly become an explanation, a structured plan may collapse into free-form narrative, and earlier constraints may lose influence. CORE prevents these shifts by supplying an explicit reasoning operator on every turn. The operator establishes the mode of transformation, while the Local Concept provides the content to which that mode applies. The model is therefore guided by a defined reasoning frame rather than by implicit reconstruction.

Together, these effects turn multi-turn dialogue from a process of repeated rediscovery into one of incremental refinement. Efficiency gains arise from bounded prompt size; stability gains arise from explicit reasoning structure and persistent task semantics. CORE achieves both without modifying the underlying model, relying instead on a restructuring of the interface through which the model is engaged.

# 4 Prototype Implementation (PoC)

To test whether a concept-first interaction layer can operate in practice, we built a small prototype implementing the CORE loop end-to-end. The PoC maintains a minimal Local Concept, selects a reasoning operator for each turn, and constructs prompts using only a compact concept summary rather than replaying prior tokens. The minimal prototype implementation can be accessed online: https://huggingface.co/spaces/vishwashegde/core-concept-pocv2

## 4.1 How it Works

Each user turn is processed in three steps. First, the system interprets the instruction and selects an operator from a small rule-based subset of the Global CORE library (e.g., summarize, refine constraints, compare, outline). Second, the system updates the Local Concept—its evolving representation of the task, constraints, and intermediate outputs—using lightweight extraction rules. Third, the next prompt to the model is constructed from (i) the operator, (ii) the Local Concept summary, and (iii) the new user instruction. No conversation history is replayed. The prototype therefore simulates the key architectural shift: continuity comes from an external semantic state rather than from token accumulation.

## 4.2 Results

Across multi-turn interactions, the PoC consistently exhibits two behaviors predicted by the CORE architecture. Prompt size stays approximately constant because the Local Concept summary does not grow with the conversation, whereas a baseline token-first prompt grows linearly. And reasoning remains more stable across turns: constraints persist, structural patterns remain coherent, and the system avoids the drift normally introduced when the model must infer context from an expanding transcript. These effects do not depend on model modification—they arise solely from changing what the model receives as input.

To illustrate the efficiency dynamics, we measured cumulative prompt-token growth across a multi-turn task under two conditions:
(1) token-first baseline, and
(2) concept-first prompting using CORE.

Table 1: Cumulative prompt-token growth across a multi-turn task. The concept-first CORE approach maintains a compact prompt and stable reasoning, while the token-first baseline grows linearly and exhibits drift.

Prompt Size per Turn (with cumulative savings)

| Turn | Baseline Prompt Tokens | CORE Prompt Tokens | Savings % | Baseline Cumulative Tokens | CORE Cumulative Tokens | Cumulative Savings % |
|---|---|---|---|---|---|---|
| 1 | 90 | 155 | -72.2 | 90 | 155 | -72.2 |
| 2 | 124 | 144 | -16.1 | 214 | 299 | -39.7 |
| 3 | 164 | 155 | 5.5 | 378 | 454 | -20.1 |
| 4 | 203 | 154 | 24.1 | 581 | 608 | -4.6 |
| 5 | 246 | 158 | 35.8 | 827 | 766 | 7.4 |
| 6 | 287 | 151 | 47.4 | 1114 | 917 | 17.7 |
| 7 | 327 | 155 | 52.6 | 1441 | 1072 | 25.6 |
| 8 | 367 | 155 | 57.8 | 1808 | 1227 | 32.1 |
| 9 | 407 | 155 | 61.9 | 2215 | 1382 | 37.6 |
| 10 | 450 | 158 | 64.9 | 2665 | 1540 | 42.2 |

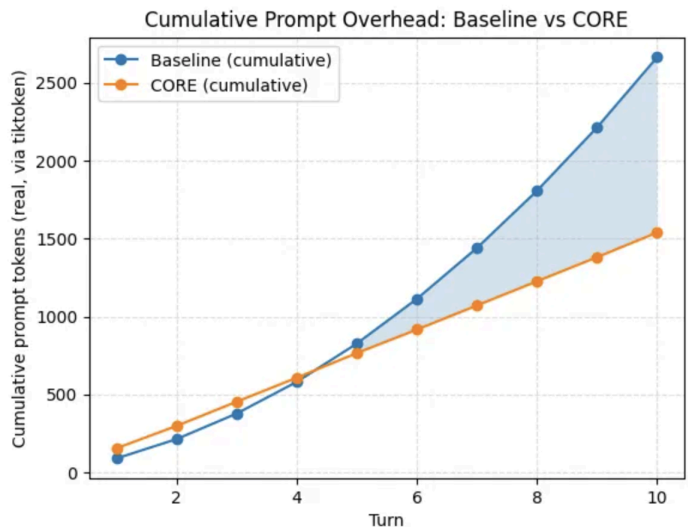

Figure 2: Visual Comparison of Baseline vs. CORE

### 4.3 Limitations

The PoC is intentionally minimal. Operator selection is rule-based rather than learned; Local Concept extraction is shallow and not domain-robust; only a single active concept is supported; and the efficiency comparison measures prompt-input size rather than full computational cost or downstream accuracy. As such, the prototype is not an empirical evaluation of CORE. Its purpose is to demonstrate that the architecture can be implemented cleanly, functions as intended in multi-turn settings, and produces the qualitative behavior the concept-first design predicts.

## 5 Discussion

CORE is presented as a conceptual and system-level hypothesis rather than a validated solution, and several structural limitations remain.

**Operator selection.** CORE depends on accurately choosing a cognitive operator each turn. Heuristic selection works for a prototype, but robust deployment likely requires learned or hybrid classifiers. Misclassification does not break the system but can yield misaligned reasoning trajectories.

**Local Concept quality.** The Local Concept must be abstract enough to remain efficient yet complete enough to prevent the model from reverting to reconstructive inference. Overly terse summaries risk ambiguity; overly detailed ones approach prompt replay. Identifying the optimal abstraction boundary is an open design challenge.

**Externalization of semantics.** CORE does not modify the model's internal representations. It constrains behavior through structured textual cues, meaning it cannot eliminate hallucinations, override model biases, or substitute for training-time alignment. It should be seen as complementary to, not a replacement for, model-level methods.

**Retention and privacy.** Maintaining compact semantic summaries across long time spans raises policy and governance questions. Although summaries contain no conversational text, deployments must determine retention windows, user consent models, and purging strategies.

**Evolution of operators and integration with other systems.** How the operator library should evolve—fixed for interpretability or adaptive to domain usage—remains open. Likewise, how Local Concepts interact with learned memory systems, agent frameworks, or symbolic planners is not yet understood.

These limitations do not undermine the architectural motivation. They indicate that CORE is an initial proposal for a concept-level interface; formal evaluation, operator learning, and deeper integration with reasoning systems are left for future work.

# 6 Conclusion

Large language models remain limited not by their expressiveness but by the structure of the interfaces through which they are used. In the prevailing token-based paradigm, models reconstruct the meaning of a task at every turn, relying on the replay of textual history and on transient internal representations that vanish after each invocation. This architecture is effective for short interactions but struggles to maintain stability, coherence, and efficiency over extended dialogues.

CORE offers an alternative framing. By separating the representation of meaning from the representation of text, and by introducing a finite grammar of cognitive operators that governs how reasoning proceeds, CORE transforms multi-turn dialogue into a stateful reasoning process. The Local Concept provides a compact, persistent account of the task that evolves across turns, while the operator library constrains the model's reasoning mode in a consistent and interpretable way. Together, these components reduce reliance on token replay, mitigate drift, and create a substrate on which more reliable multi-step interactions can be built.

The architecture does not modify the underlying model, nor does it claim to solve the broader challenges of alignment or factual accuracy. Instead, it proposes a system-level approach that complements existing methods. The prototype implementation presented here is preliminary, but it illustrates a direction in which interface-layer design may meaningfully improve the usability of large models, especially in applications that require sustained reasoning, revision, and contextual continuity.

The broader implication is conceptual: language models need not operate exclusively within the constraints of token histories. By externalizing meaning and introducing explicit structure into the reasoning process, it becomes possible to build systems in which models function not only as generators of text but as components in a larger architecture capable of stable, interpretable, multi-turn interaction. The development of such architectures invites further investigation and may represent a productive path forward for the next generation of language model systems.

# References

Bang, Y., Cahyawijaya, S., Lee, N., Dai, W., Su, D., Wilie, B., Lovenia, H., Ji, Z., Yu, T., Chung, W., Do, Q. V., Xu, Y., & Fung, P. (2023). A multitask, multilingual, multimodal evaluation of ChatGPT on reasoning, hallucination, and interactivity. ArXiv:2302.04023 [Cs]. https://arxiv.org/abs/2302.04023

Bender, E. M., Gebru, T., McMillan-Major, A., & Shmitchell, S. (2021). On the dangers of stochastic parrots: Can language models be too big? Proceedings of FAccT '21, 610–623. https://doi.org/10.1145/3442188.3445922

Lewis, P., Perez, E., Piktus, A., Petroni, F., Karpukhin, V., Goyal, N., Küttler, H., Lewis, M., Yih, W., Rocktäschel, T., Riedel, S., & Kiela, D. (2021, April 12). Retrieval-augmented generation for knowledge-intensive NLP tasks. ArXiv.org. https://doi.org/10.48550/arXiv.2005.11401

Lightfoot, C., Cole, M., & Cole, S. R. (2018). The development of children (8th ed.). Worth Publishers / Macmillan Education.

Ouyang, L., Wu, J., Jiang, X., Almeida, D., Wainwright, C. L., Mishkin, P., Zhang, C., Agarwal, S., Slama, K., Ray, A., Schulman, J., Hilton, J., Kelton, F., Miller, L., Simens, M., Askell, A., Welinder, P., Christiano, P., Leike, J., & Lowe, R. (2022). Training language models to follow instructions with human feedback. ArXiv:2203.02155 [Cs]. https://arxiv.org/abs/2203.02155

Piaget, J., & Inhelder, B. (1969). The psychology of the child. Basic Books.

Press, O., Zhang, M., Min, S., Schmidt, L., Smith, N. A., & Lewis, M. (2023, May 22). Measuring and narrowing the compositionality gap in language models. ArXiv.org. https://doi.org/10.48550/arXiv.2210.03350

Schick, T., Dwivedi-Yu, J., Dessì, R., Raileanu, R., Lomeli, M., Zettlemoyer, L., Cancedda, N., & Scialom, T. (2023). Toolformer: Language models can teach themselves to use tools. ArXiv:2302.04761 [Cs]. https://arxiv.org/abs/2302.04761

Vaswani, A., Shazeer, N., Parmar, N., Uszkoreit, J., Jones, L., Gomez, A. N., Kaiser, L., & Polosukhin, I. (2017, June 12). Attention is all you need. Cornell University. https://arxiv.org/abs/1706.03762

Wei, J., Wang, X., Schuurmans, D., Bosma, M., Ichter, B., Xia, F., Chi, E., Le, Q., & Zhou, D. (2022). Chain of thought prompting elicits reasoning in large language models. ArXiv:2201.11903 [Cs]. https://arxiv.org/abs/2201.11903

Yao, L., Mao, C., & Luo, Y. (2019). KG-BERT: BERT for knowledge graph completion. ArXiv.org. https://arxiv.org/abs/1909.03193

## Acknowledgements

The authors used ChatGPT solely for language editing and clarity improvements. All technical ideas, architectural decisions, analyses, and conclusions were developed and verified by the authors.

## Disclosure

# Appendix 1
## Global CORE Operator Library

| Family | Operator | Function |
| --- | --- | --- |
| Distillation & Extraction | Summarize | Condense content |
| Distillation & Extraction | Abstract | Move from specifics to generalities |
| Distillation & Extraction | Extract Facts | Identify verifiable statements |
| Distillation & Extraction | Extract Entities | Identify key items or variables |
| Distillation & Extraction | Highlight Constraints | Surface requirements or limits |
| Distillation & Extraction | Highlight Risks | Identify uncertainties or failure points |
| Distillation & Extraction | Generalize Pattern | Identify recurring structure |
| Distillation & Extraction | Extract Relationships | Identify links between entities or facts |
| Transformation & Rewriting | Rewrite (Tone) | Change tone while preserving meaning |
| Transformation & Rewriting | Rewrite (Style) | Change structure or format |
| Transformation & Rewriting | Simplify | Make accessible to non-experts |
| Transformation & Rewriting | Elaborate | Expand with richer detail |
| Transformation & Rewriting | Paraphrase | Restate without altering intent |
| Transformation & Rewriting | Translate | Convert between languages |
| Transformation & Rewriting | Convert Format | Shift between tables, bullets, prose |
| Transformation & Rewriting | Normalize Language | Standardize style and remove slang |
| Explanatory Reasoning | Explain | Clarify directly |
| Explanatory Reasoning | Explain by Analogy | Use analogy to convey structure |
| Explanatory Reasoning | Define | Provide concise definition |
| Explanatory Reasoning | Resolve Ambiguity | Remove multiple interpretations |
| Explanatory Reasoning | Trace Cause–Effect | Show causal or logical links |
| Explanatory Reasoning | Fill Missing Steps | Add intermediate reasoning |
| Explanatory Reasoning | Contrast Concepts | Highlight key differences |
| Explanatory Reasoning | Provide Rationale | Explain reasons behind a claim or action |
| Planning & Structuring | Outline | Produce structured plan |
| Planning & Structuring | Decompose | Break into parts or subtasks |
| Planning & Structuring | Plan Steps | Order actions toward a goal |
| Planning & Structuring | Prioritize | Rank by criteria |
| Planning & Structuring | Sequence | Determine correct or efficient order |
| Planning & Structuring | Generate Variants | Produce meaningful alternatives |
| Planning & Structuring | Map Dependencies | Identify task or concept dependencies |

| Planning & Structuring | Estimate Effort | Provide coarse effort/complexity estimates |
| Evaluation & Verification | Evaluate Quality | Judge clarity or correctness |
| Evaluation & Verification | Rank/Score | Provide ordinal or quantitative score |
| Evaluation & Verification | Check Consistency | Ensure internal coherence |
| Evaluation & Verification | Check Constraints | Verify requirement adherence |
| Evaluation & Verification | Groundedness Check | Ensure claims follow input |
| Evaluation & Verification | Reflect & Restate | Summarize current state |
| Evaluation & Verification | Detect Contradictions | Identify conflicting statements |
| Evaluation & Verification | Validate Assumptions | Surface unstated assumptions |

# Appendix 2

## Operator Structure (Compact Representation)

Each Global CORE operator is represented as a structured specification that defines the reasoning transformation to be performed. All operators conform to the following schema:

```
{
 "operator_id": "COMPARE",
 "description": "Produce a structured comparison between two or more entities.",
 "input_requirements": ["intermediate_results must contain >= 2 items"],
 "expected_output": "Parallel comparison structure or table",
 "reasoning_constraints": [
   "Use only constraints stored in Local Concept",
   "Do not introduce new evaluation criteria"
 ],
 "state_update_rules": [
   "Store comparison summary in intermediate_results",
   "Mark resolved attributes as clarified"
 ]
}
```

All operators in Appendix 1 follow this format, differing only in their descriptions, constraints, and expected outputs.

# Appendix 3

## Illustrative Multi-Turn Examples

The following examples provide high-level demonstrations of how CORE manages meaning across turns through operator selection and concept-state evolution. These are not prompts; they illustrate semantic transitions.

### 3.1 Deepening a Topic: From Initial Query to Structured Comparison

**Turn 1 — User Input**

"What are some good dog breeds for small children?"

**Operator:** Summarize → Generate Candidates
 **Local Concept (after update):**
 Task: select suitable dog breed for family with small children
 Constraints: none
 Intermediate: shortlisted breeds = {Beagle, Labrador Retriever, Poodle}

**Turn 2 — User Input**

"We live in an apartment, and shedding is a concern."

**Operator:** Update Constraints
 **Local Concept:**
 Task: same
 Constraints: apartment-friendly, low shedding
 Intermediate: refined shortlist = {Poodle, Miniature Schnauzer}

**Turn 3 — User Input**

"Compare those two."

**Operator:** Compare
 **Local Concept:**
 Task: select dog breed
 Constraints: apartment-friendly, low shedding
 Intermediate: breeds to compare = {Poodle, Miniature Schnauzer}

The model generates the comparison using a compact concept summary rather than replaying all turns.

### 3.2 Topic Shift and Return

**Turn X — New Topic Introduced**

"Switching gears—can you explain quantum entanglement?"

**Operator:** Explain
 **Local Concept A (dog task):** moves to dormant state
 **Local Concept B (physics task):** becomes active

**Turn X+N — User Returns**

"By the way, which breed did we shortlist earlier?"

CORE reactivates Local Concept A:
 Task: dog breed selection
 Constraints: apartment-friendly, low shedding
 Intermediate: shortlist = {Poodle, Miniature Schnauzer}

The system resumes seamlessly without reconstructing from tokens.

### 3.3 Ambiguity Resolution in Multi-Step Tasks

**Turn 1**

"I need help creating a business plan."

**Operator:** Outline
 **Local Concept:**
 Task: create business plan outline
 Constraints: none
 Intermediate: draft outline sections

**Turn 2**

"Make the financial section more detailed."

**Operator:** Elaborate
 **Local Concept:**
 Task unchanged
 Intermediate: financial section now contains expanded subsections
 Pending: validation of assumptions

**Turn 3**

"These projections seem too optimistic."
 **Operator:** Evaluate
 CORE updates intermediate results and constraints under a stable meaning representation.